\documentclass[runningheads]{llncs}
\usepackage[T1]{fontenc}
\usepackage{graphicx}
\usepackage{multirow}%
\usepackage{changepage}
\usepackage{amsmath,amssymb,amsfonts}%
\usepackage{mathrsfs}%
\usepackage[title]{appendix}%
\usepackage{xcolor}%
\usepackage{textcomp}%
\usepackage{manyfoot}%
\usepackage{booktabs}%
\usepackage{algorithm}%
\usepackage{algorithmicx}%
\usepackage{algpseudocode}%
\usepackage{listings}%
\usepackage{adjustbox}
\usepackage{subcaption}
\usepackage{tikz}
\usepackage{wrapfig}
\usepackage{dblfloatfix}

\begin{document}
\title{RevCD - Reversed Conditional Diffusion for \\ Generalized Zero-Shot Learning}
\titlerunning{RevCD}
%

\author{
William Heyden              \orcidID{0009-0004-1902-2135} \and
Habib Ullah                 \orcidID{0000-0002-2434-0849} \and
Muhammad Salman Siddiqui    \orcidID{0000-0002-6003-5286} \and
Fadi Al Machot              \orcidID{0000-0002-1239-9261}
}
\authorrunning{W. Heyden et al.}
\institute{Norwegian University of Life Sciences, \\ 1433 Ås, Norway \\
\email{\{william.hedyen, habib.ullah, muhammad.salman.siddiqui, fadi.al.machot\}@nmbu.no}
}

\maketitle 
\begin{abstract}
    In Generalized Zero-Shot Learning (GZSL), we aim to recognize both seen and unseen categories using a model trained only on seen categories. In computer vision, this translates into a classification problem, where knowledge from seen categories is transferred to unseen ones by exploiting the relationships between visual features and available semantic information. However, learning this joint distribution is costly and requires one-to-one alignment with corresponding semantic information. We present a reversed conditional diffusion-based model (RevCD) that mitigates this issue by estimating the semantic density conditioned on visual inputs. Our RevCD model consists of a cross Hadamard-addition embedding of a sinusoidal time schedule, and a multi-headed visual transformer for attention-guided embeddings. The proposed approach introduces two key innovations. First, we apply diffusion models to zero-shot learning, a novel approach that exploits their strengths in capturing data complexity. Second, we reverse the process by approximating the semantic densities based on visual data, made possible through the classifier-free guidance of diffusion models. Empirical results demonstrate that RevCD achieves competitive performance compared to state-of-the-art generative methods on standard GZSL benchmarks. The complete code will be available on GitHub.
\keywords{Zero-shot Learning  \and Transfer Learning \and Diffusion Model}
\end{abstract}
\section{Introduction}
\label{sec:intro}
Zero-shot learning (ZSL) represents a state-of-the-art advancement in machine learning transferability and computer vision classification. By pushing the boundaries of knowledge extraction, ZSL enable ML models to expand without costly retraining. This learning paradigm is particularly crucial as it addresses the inherent limitation of traditional machine learning models that require prior access to expensive datasets. ZSL leverages auxiliary knowledge, allowing models to explore unobserved events, edge cases, or new compositions without any additional training. Traditional approaches in ZSL focused on aligning attributes directly with object categories \cite{lampert2009learning}, 
while deep learning's potential to create a joint embedding space of visual and semantic features \cite{chen2020rethinking,mishra2018generative,chen2021semantics,frome2013devise,socher2013zero} has rendered this approach obsolete. The shift toward latent-based methods highlights the importance of embedding space techniques because of their ability to decode and infer complex data distributions. This offers a promising resolution to the two main challenges of ZSL: the semantic gap and limited generalization ability \cite{verma2017simple}.

\begin{figure}[t]
    \centering
    \includegraphics[scale=0.40]{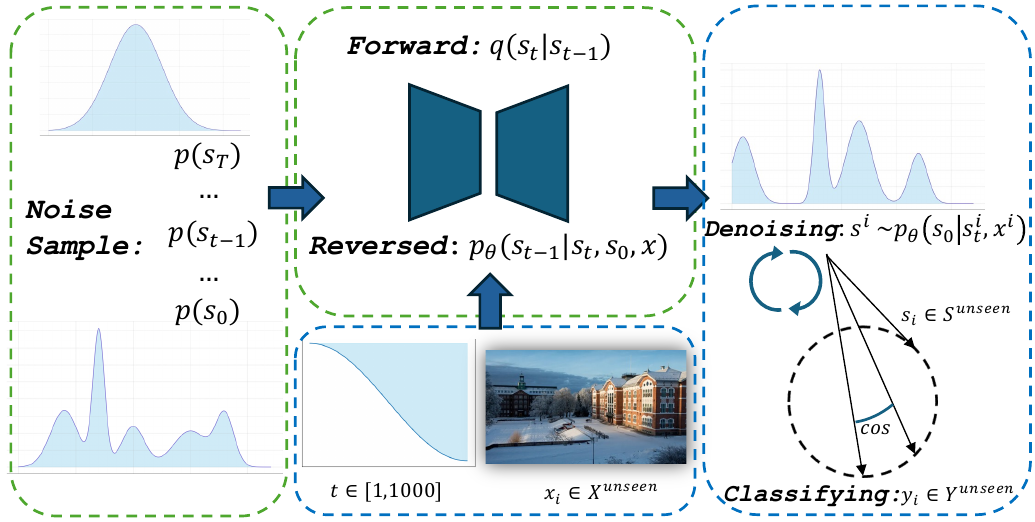}
    \caption{\small{Overview of the proposed RevCD model: We train a denoising process using only seen samples (indicated by green boxes). Once trained, the model can estimate the semantic distribution by conditioning on the visual space of unseen samples (represented by blue boxes) and Gaussian noise. The final classification is conducted through a simple nearest-neighbor search based on the estimated density.}}
    \label{fig:overview}
\end{figure}

Our contribution introduces a Diffusion-based generative model, a notable innovation in ZSL. Distinct from conventional models that predominantly rely on attribute matching or embedding strategies, our RevCD model utilizes a diffusion process to model the data distributions iteratively, see Fig. \ref{fig:overview}. This augments the model's capability to manage class variability and enhances its generalization capacity. Such control over the semantic space is required for overcoming the challenges of bias and hubness commonly encountered in ZSL methodologies \cite{lazaridou2015hubness}.

\section{Related works}

Advancements in likelihood-based models have been central to the progress of zero-shot learning \cite{pourpanah2022review}. By framing the learning process as a maximum likelihood estimation problem, these methods effectively model data distributions, allowing for robust generalization across both seen and unseen classes. This section categorizes ZSL approaches according to the foundational models employed for approximation of the data, including Variational Autoencoders (VAEs), Generative Adversarial Networks (GANs), and Hybrid models. Additionally, it highlights the role of attention mechanisms and embedding strategies in enhancing these models' performance.

\textbf{VAE-based.}
Variational Autoencoders \cite{kingma2013auto} play a crucial role in ZSL due to their probabilistic framework for modeling latent spaces. Their adaptability in synthesizing unseen class prototypes, as demonstrated by \cite{schonfeld2019generalized,bucher2017generating}, and \cite{ji2023zero}, underscores their versatility in ZSL applications. For instance, \cite{heyden2023integral} incorporates a semantic-guided approach within a VAE framework, while \cite{wang2023generalized} employs a decoupling strategy to enhance performance. A significant advantage of VAEs lies in their capacity to \textit{explicitly} approximate data density. However, a key limitation is their tendency to generate blurry or overly smoothed features due to posterior collapse \cite{lucas2019don}, which can obscure essential class-specific details needed to distinguish between unseen classes.

\textbf{GAN-based.}
Generative Adversarial Networks \cite{goodfellow2014generative} offer a powerful and dynamic framework for feature synthesis. GANs have been successfully adapted to generate features for unseen classes \cite{gao2020zero,huang2019generative,gupta2023generative,zhang2023data}. GANs´ ability to produce sharp and realistic features through \textit{implicit} density estimation makes them particularly effective for capturing fine-grained details. However, they are also prone to challenges such as training instability and mode collapse \cite{bau2019seeing}, which can result in a lack of diversity in the generated features. This limitation may hinder the model’s ability to accurately represent the full spectrum of unseen classes.

\textbf{Hybrids.}
Hybrid models in ZSL leverage multiple architectures to enhance performance. The majority of hybrid frameworks incorporate sequential modules, as seen in \cite{han2021contrastive,ding2023enhanced,liu2024transductive}. Employing a VAE to learn an embedding function that constrains the semantic or visual space allows for greater control over the generation of synthesized features. Nevertheless, the complexity and computational overhead of combining multiple models can pose challenges, especially in terms of model interpretability and scalability.

\textbf{Attention and embedding.}
Attention mechanisms \cite{hao2023learning,khan2023learning} and embedding strategies \cite{akata2015label,xian2016latent} further refine the latent space by focusing on salient attributes and mapping visual data to semantic space. The approaches \cite{wang2023zero,alamri2023implicit,li2023distilled} enhance interpretability and feature distinctness; however, they rely heavily on high-quality, granular attribute information, which is not always available, limiting their applicability across diverse datasets.


Our contribution introduces a reversed Diffusion-based model (RevCD) for zero-shot inference. Diffusion models have been applied to improve accuracy as generative classifiers \cite{chen2023robust,azizi2023synthetic,shipard2023diversity}, and their capacity to generate synthetic data has enabled the classification of unseen compositions \cite{clark2024text,li2023your}. However, existing applications are limited by pre-training on prompt categories. Their implementation in a pure zero-shot setting is still absent. We address these limitations by leveraging the reversed process for generating conditioned semantic embeddings, aiming for effective generalization to unseen classes without the constraints observed in the aforementioned methodologies.  \textit{To the best of our knowledge, Diffusion models have not yet been explored within the ZSL domain.}

\begin{figure*}[t]
    \centering
    \begin{subfigure}[t]{0.55\textwidth}
        \centering
        \includegraphics[width=\textwidth]{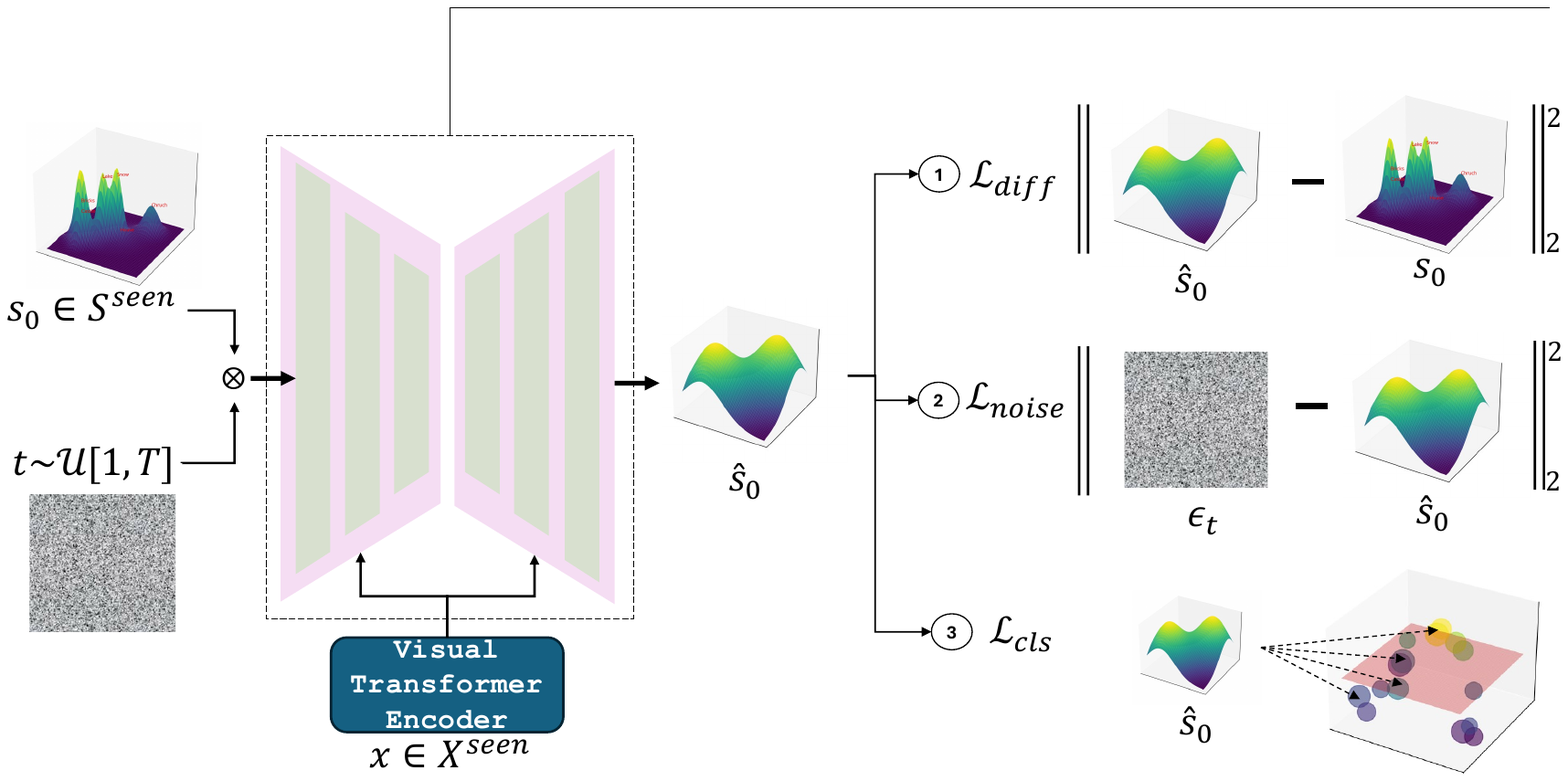}
        \caption{}
        \label{fig:first}
    \end{subfigure}
    \begin{tikzpicture}
        \draw[gray, thin] (0,0) -- (0,-6);
    \end{tikzpicture}
    \begin{subfigure}[t]{0.35\textwidth}
        \centering
        \includegraphics[width=\textwidth]{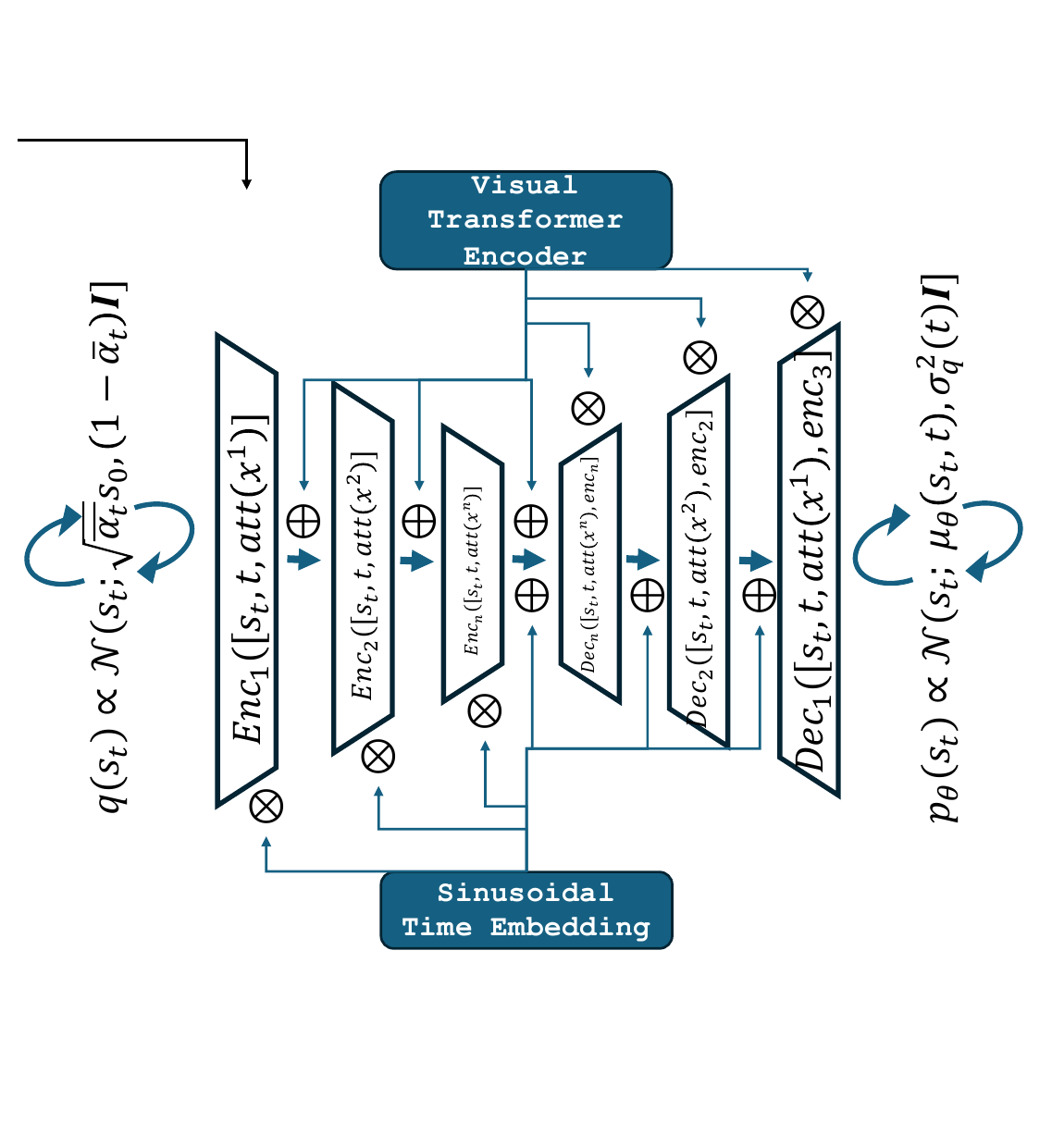}
        \caption{}
        \label{fig:second}
    \end{subfigure}
    \caption{The figure illustrates our proposed approach for training. \textbf{(a)} presents our high-level architecture and associated loss functions. By conditioning the image, we can infer the semantic distribution of unseen classes. \textbf{(b)}  provides a detailed view of our U-net architecture. It implements sinusoidal time and cross-Hadamard-Addition conditional embeddings for optimal control over the learned distribution. In ZSL, the goal is to transfer the knowledge of how to infer the distribution rather than the distribution itself.}
    \label{fig:combined}
\end{figure*}

\section{Methodology}
\subsubsection{Problem Setup.}
We denote the set of seen images, semantics, and corresponding class labels as \(\{x^{{seen}}, s^{{seen}}, y^{{seen}}\} \in \mathcal{D}^{{seen}}\), where \(x^{{seen}}\) represents the images, \(s^{{seen}}\) the semantics, and \(y^{{seen}}\) the class labels for the seen classes. The set of unseen semantics and class labels, denoted as \(\{s^{{unseen}}, y^{{unseen}}\} \in \mathcal{D}^{{unseen}}\), represents the unseen dataset. During training, the model is trained exclusively on data from the seen set \(\mathcal{D}^{{seen}}\), while assuming access to the semantic and label information of the unseen classes in \(\mathcal{D}^{{unseen}}\). Importantly, the unseen images, \(x^{{unseen}}\), are \textit{not} available during training and are only introduced during the inference phase. It is important to note that during training, the set of class labels for seen and unseen data does not overlap, i.e., \(\mathcal{Y}^{seen} \cap \mathcal{Y}^{unseen} = \emptyset\). This ensures there is no direct overlap between seen and unseen classes. During inference, the challenge is to map an unseen sample image, \(x^{unseen}\), to its corresponding unseen label \(y^{unseen}\), using a learned function \(f: x^{unseen} \rightarrow y^{unseen}\). The training process involves using paired examples \(\{x^{seen}, s^{{seen}}\} \in \mathcal{D}^{{seen}}\) to learn a joint model, $p_{\theta}(s,x)$, that can be used to estimate the conditional probability distribution function for the semantic space \(s^{{seen}}\) given their corresponding visual features \(x^{{seen}}\) as $p_{\theta}(s|x^{seen})$. During the test phase, the semantic distribution of the unseen images \(x^{{unseen}}\) is approximated and subsequently classified into their corresponding unseen classes \(y^{{unseen}} \in \mathcal{Y}^{{unseen}}\). 

\subsection{Diffusion Process}
The diffusion process \cite{sohl2015deep} models complex data distributions through a specific Markov chain structure. During training, we start with time-step $0$ represented as an ordinary, clean, sample from the semantic space $s_0 \sim q(S)$. This is paired with its corresponding visual features $\{s_0, x\} \in \mathcal{D}^{seen}$. We incrementally infuse Gaussian noise using a fixed linear Gaussian model $s_1, \dots, s_T$, for $T$ steps. By employing the reparameterization trick \cite{kingma2013auto}, we can parameterize the mean $\mu_t(s_t) = \sqrt{\alpha_t}s_{t-1}$, and variance $\Sigma(s_t) = (1-\alpha_t)\mathbf{I}$ for hierarchical time-steps $t \in [0,T]$. Pre-defining the noise schedule $(\beta_1, \dots, \beta_T)$ allows us to sample from the Markov chain through a fixed forward sequence of time steps as:

\begin{equation}
    q(s_t|s_{t-1}) = N(s_t; \sqrt{\alpha_t}s_{t-1}, (1-\alpha_t)\bold{I})
\end{equation}

where $\alpha_t = 1 - \beta_t$ and $\bar{\alpha}_t = \prod_{i=1}^{t} \alpha_i$. This forward encoding process has the desired properties of being variance-preserving and completely deterministic, and the final distribution $p(s_T)$ is a standard Gaussian. Our aim is to learn a reverse diffusion kernel $p(s_{t-1}| s_t, x)$ which removes the noise of the forward process and can estimate a clean sample, $\hat{s}_0$, from random noise conditioned on a visual space:

\begin{equation}
    p(s_{0:T}|x) = p(s_T)\prod_{t=1}^{T}p(s_{t-1}|s_t,x) 
\end{equation}

Directly expressing $p(\cdot)$ in closed form is intractable. We instead parameterize $p_{\theta}$ with $\theta$ and approximate the distribution by minimizing the evidence lower bound (ELBO) of the conditional log-likelihood

\begin{equation}
    log\ p_{\theta}(s|x) \geq \mathbb{E}_{q(s_{1:T}|s_0,x)} \left[\log \frac{p_{\theta}(s_0|s_{1:T},x)p(s_{1:T},x)}{q(s_{1:T}|s_0, x)} \right]
\end{equation}

\begin{equation}
    \label{eq:elbo}
    \begin{split}
    = \underbrace{\mathbb{E}_{q(s_1, x|s_0)}[\log p_{\theta}(s_0|s_1, x)]}_{\text{reconstruction}} - \underbrace{D_{KL}[q(s_T|s_0, x)||p(s_T|x)]}_{\text{prior matching}} 
    \\ - \underbrace{\sum^T_{t=2} \mathbb{E}_{q(s_t, x|s_0)}[D_{KL}[q(s_{t-1}|s_t, s_0, x)|| p_{\theta}(s_{t-1}|s_t, x)]]}_{\text{diffusion term}}
    \end{split}
\end{equation}


By conditioning the forward process on the clean example at any given $t$, the diffusion loss can be formulated using Bayes´ rule as the KL divergence between the ground-truth analytical denoising step $q(s_{t-1}|s_t, s_0, \bold{x})$ and our approximated denoising step $p_{\theta}(s_{t-1}|s_t, \bold{x})$. The prior loss $D_{KL}[q(s_T|s_0, x)||p(s_T|x)]$ can be ignored as it does not contain any trainable parameters and is zero under our assumption. The reconstruction loss $\log p_{\theta}(s_0|s_1, x)$ is typically minimal and can be safely ignored without affecting the outcome. Therefore, our diffusion objective becomes:

\begin{equation}\label{eq:objective}
    \operatorname*{arg\,max}_\theta \mathbb{E}_{t \sim [2,T]}[\mathbb{E}_q[D_{KL}(q(s_{t-1}|s_t, s_0)||p_{\theta}(s_{t-1}|s_t, x))]]
\end{equation}

which boils down to learning a neural network, $s_{\theta}$, to predict the semantic space $\hat{s}_0$ from noise at time $t$, conditioned on an image $x$. This network can be optimized using stochastic samples of $t$ from a uniform distribution $t \sim \mathcal{U}[1,T]$. 

Given our case, where we can set the variances to match exactly, the KL divergence in Eq. (\ref{eq:objective}) can be reduced to a minimization of the difference between the mean of the two distributions \cite{duchi2007derivations}. 

\begin{equation}
    \operatorname*{arg\,min}_{\theta} D_{KL}(q(s_{t-1}|s_t,s_0)||p_{\theta}(s_{t-1}|s_t, x))
\end{equation}
\begin{equation}
    \label{eq:mu_diff}
    = \operatorname*{arg\,min}_{\theta} \frac{1}{2\sigma_q^2(t)}[||\mu_{\theta}(\cdot)-\mu_q(\cdot)||^2_2]
\end{equation}

where $\mu_q$ is the ground truth transition mean predetermined in the forward process and $\mu_{\theta}$ is the mean of our learned diffusion kernel.

\subsubsection{Reconstruction loss.}
Using Bayes' rule and the Markov assumption, $q(s_t|s_{t-1}, s_0) = q(s_t|s_{t-1})$ \cite{lu2022learn}, we can show that:
\begin{equation}
    q(s_{t-1}|s_t, s_0) \propto \mathcal{N}(s_{t-1}; 
    \frac{\sqrt{\alpha_t}(1 - \bar{\alpha}_{t-1})s_t + \sqrt{\bar{\alpha}_{t-1}}(1 - \alpha_t)s_0}{1 - \bar{\alpha}_t},\underbrace{\frac{(1 - \alpha_t)(1 - \bar{\alpha}_{t-1})}{1 - \bar{\alpha}_t} \mathbf{I}}_{\sigma_q^2(t)})
\end{equation}

Employing this relationship to eq. (\ref{eq:mu_diff}), we arrive at our formulation of diffusion loss (loss \raisebox{.5pt}{\textcircled{\raisebox{-.9pt} {1}}} in Fig. \ref{fig:first}), to predict the ground truth sample from an arbitrary noised version of it:
\begin{equation}
    \mu_q(\mathbf{s}_t; s_0) = \frac{\sqrt{\alpha_t}(1 - \bar{\alpha}_{t-1})s_t + \sqrt{\bar{\alpha}_{t-1}}(1 - \alpha_t)s_0} {1 - \bar{\alpha}_t}
\end{equation}

\begin{equation}\label{eq:diffusion}
    \mathcal{L}_{reconstruction} = \underbrace{\frac{1}{2\sigma_q^2(t)} \frac{\bar{\alpha}_{t-1}(1-\alpha_{t})^2}{(1-\bar{\alpha}_t)^2}}_{w_t} \left[||s_0-s_{\theta}(s_t, t, x)||_2^2\right]
\end{equation}

Here, we replace $\mu_q(\cdot) = s_0$ with $\mu_{\theta}(\cdot)=s_{\theta}(s_t,t,x)$ to estimate the reversed diffusion kernel. The first term, $w_t$, is a time-dependent variance weight, where $\sigma_q^t(t) = \beta_t$ \cite{ho2020denoising}. However, empirical research \cite{li2023your} has demonstrated that setting $w_t=1$ yields optimal performance. Our experiments in the zero-shot paradigm show consistent outcomes.

\subsubsection{Noise loss.}
The diffusion loss in Eq. (\ref{eq:mu_diff}) can also be interpreted as estimating the source noise added $\hat{\epsilon}_t$, rather than directly predicting the clean sample $\hat{s}_0$. By applying the reparameterization trick \cite{luo2022understanding}, we can express the relationship between a clean and an arbitrarily noised sample as:

\begin{equation}
    s_0 = \frac{s_t-\sqrt{1-\bar{\alpha}_t}\epsilon_0}{\sqrt{\bar{\alpha}_t}}
\end{equation}

This enables us to estimate the reverse transition mean by directly utilizing the estimated added noise instead:

\begin{equation}
    \mu_{q}(s_t;s_0) = \frac{1}{\sqrt{\alpha_t}}s_t - \frac{1-\alpha_t}{\sqrt{1-\bar{\alpha}_t}\sqrt{\alpha_t}}\epsilon_0
\end{equation}

By incorporating this reformulation into Eq. (\ref{eq:mu_diff}), as a function of the perturbed noise at time step $t$, we optimized the network by estimating the source noise from the predicted noise. Leveraging this theoretical perspective, we introduce an additional noise loss term that corresponds to our clean-sample predictor (loss \raisebox{.5pt}{\textcircled{\raisebox{-.9pt} {2}}} in Fig. \ref{fig:first}):

\begin{equation}
    \mathcal{L}_{noise} = \underbrace{\frac{1}{2\sigma_q^2(t)} \frac{(1-\alpha_t)^2}{(1-\bar{\alpha}_t)^2\alpha_t}}_{w'_t}[||\epsilon_0 - \frac{\sqrt{\bar{\alpha}_t}}{\sqrt{1-\bar{\alpha}_t}s_t}s_{\theta}(s_t,t,x)||_2^2]
\end{equation}

Note the slight variance-weighting difference compared to the diffusion loss in Eq. (\ref{eq:diffusion}). This discrepancy is a correction term as a result of the different transition mean calculation. However, it can be eliminated by assigning a fixed constant value of \(w'_t = 1\).

These two complementary formulations of the denoising transition mean correspond to an equivalent optimization problem (Eq. \ref{eq:objective}). Although these formulations introduce additional complexity to the optimization process, necessitating more sophisticated strategies and careful hyperparameter tuning to achieve convergence, we observe significant improvements in density estimation. We believe that this loss function acts as a regularizer during optimization, since $w'_t \neq w_t$, enhancing the model’s ability to navigate the loss landscape and generalize to unseen data. By optimizing from multiple perspectives, the model generates richer and more robust representations.

\subsubsection{Classification loss.}
Unlike other generative models such as flow-based models and GANs, Diffusion models have no natural property to decrease the intra-class variance from the noise input \cite{ho2022classifier}. Previous work in classifier- and classifier-free guidance in score-based Diffusion models \cite{ho2022classifier} involves modifying the score function with the gradient of the log-likelihood of a separate classifier model $\vartheta$, $-\nabla_{\vartheta} \log p_\vartheta(y|s_t)$. Our classification objective is to steer our optimization problem of the inferred distribution through manifold regularization \cite{wang2015semi}, leveraging a classifier:
\begin{equation}
    \operatorname*{arg\,max}_\vartheta \mathbb{E}_{y \in Y_{seen}} [\log \ p_{\vartheta}(\hat{s}_{0:T} | y)]
\end{equation}

This allows us to approximate samples from the distribution $p_{\theta}(y|s_t) \propto p_{\theta}(s_t|y)p_{\theta}(y)$.
The strategy of assigning higher likelihood to the correct label has led to notable improvements in both the perceptual qualities and the inception scores of models, as highlighted in prior research \cite{salimans2016improved}. However, within the zero-shot learning framework, our goal shifts towards enhancing the model's ability to generalize the learned distribution for generating samples. These samples are not primarily focused on visual appeal but are aimed at positioning the probability mass of each conditional sample at a greater distance. Therefore, we formulate the loss as the expectation over the empirical sample distribution $\mathbb{E}[\mathcal{L}(f(\bold{x};\vartheta),y_x)]$ and implement this with a cross-entropy loss (loss \raisebox{.5pt}{\textcircled{\raisebox{-.9pt} {3}}} in Fig. \ref{fig:first}):

\begin{equation}
    \mathcal{L}_{classification} = - \frac{1}{n} \sum_{i=1}^{n} \sum_{j=1}^{c} y_{ij} \log p_{\vartheta}(\hat{y}_{ij}|s_i)
\end{equation}
where $n$ is the number of samples and $y_{ij}$ and $\hat{y}_{ij}$ are the true and predicted label for class $j$ of the $i$-th instance.

\subsection{Training Objective} \label{sec:training_objective}
Our main idea focuses on directly modeling the semantic posterior using variational inference rooted in Eq. (\ref{eq:total_loss}). We achieve this by disentangling the posterior estimation into three key components: noise prediction, data reconstruction, and (auxiliary) classification. This decomposition results in a more complex and nuanced loss landscape \cite{li2018visualizing}. Despite the increased complexity, integrating these distinct loss components enhances the model’s generalization capabilities. This is primarily due to the regularization effects inherent in the multi-faceted loss function and the fine-tuning achieved through careful hyperparameter optimization.
To enable classifier-free guidance at inference, we adopt a conditional dropout strategy during training, randomly masking the visual condition when computing the noise prediction loss \cite{ho2022classifier}. This allows the model to learn both conditional and unconditional score estimates within a single unified network. Our overall training objective becomes: 

\begin{equation}\label{eq:total_loss}
\begin{split}
        \mathbf{\mathcal{L}_{total}} = \\
    \lambda_1\mathcal{L}_{rec} + \lambda_2\mathcal{L}_{noise} + \lambda_3\mathcal{L}_{classification}
\end{split}
\end{equation}

Here, \( \lambda_1 \) and \(\lambda_3\) serve as a balancing factors between the objectives of reconstruction and classification, while \(\lambda_2\) acts as a regularization coefficient. Through this, the probability distribution of the samples aligns with the expectation of the generated conditional samples $p(s) \propto \mathbb{E}_{x\sim p(x)} [p_{\theta}(s|x)]$. The implementation details of this loss function during training are provided in Algorithm (\ref{alg:train}).

\subsection{Architectural considerations}
\subsubsection{Cross Hadamard-Addition Embeddings.}
\label{cross-had}
In the traditional diffusion process, we predict the ground truth of a noisy sample at time $t$. In our approach, however, we further condition this process on visual features. Consequently, the neural network $s_{\theta}$ is trained on the triplet $(s_t, t, x)$, where  $\{s, x\} \in \mathcal{D}^{seen}$. To refine the embeddings for both the conditioning variable $x$ and the time-step schedule $t$, we employ a cross Hadamard-Addition method, which enhances the representation and integration of these features within the network. During the representational mapping stage within the network, we use Hadamard integration for the time-step input, acknowledging that the added noise is entirely deterministic, while integrating the visual condition through addition (refer to the encoding step in Fig. \ref{fig:second}). In contrast, during the network’s generative stage, we reverse these roles, applying Hadamard integration for a stronger conditional reconstruction and a more relaxed incorporation of the time-step input (see the decoding step in Fig. \ref{fig:second}). We observe that this approach results in a closer alignment of the joint probability space, leading to improved accuracy.

\subsubsection{Time-dependent embedding.}
To increase the dimension of the time step, $t$, we employ a sinusoidal time embedding $\bar{t} \leftarrow TE(t,d)$:

\begin{equation}
\begin{split}
        \text{TE}(t, d) = [\cos(t \cdot f_0), \sin(t \cdot f_0), \ldots, \\ \cos(t \cdot f_{\frac{d}{2}-1}), \sin(t \cdot f_{\frac{d}{2}-1})],
\end{split}
\end{equation}

where \( d \) is the embedding dimension and \( f_i \) are frequencies. The temporal encoding dimension is matched to each layer of the network to learn the denoising function, given any timestep $t$. 

\subsubsection{Visual-dependent embedding.}
We implement a Transformer encoder \cite{vaswani2017attention} to extract visual features for conditioning. These visual features are integrated into the network at each layer, with the Transformer trained concurrently. To align the denoising feature dimensions, we map the multi-head attention outputs from the visual space to each intermediate feature using a Hadamard product in the decoder and matrix addition in the encoder of our denoising model (see Fig. \ref{fig:second}).

\subsubsection{Pre-conditioning.}
We also perform an affine transformation of our semantic space $s \in [0,1]$ to resemble a zero-mean Gaussian $s\prime = 2 \times s-1 \in [-1,1]$. This increases the dynamic range leading to better gradient flow and stabilizes convergence as the variance $Var(s_0) \ll 1$ skews the signal-to-noise ratio when scaled by the noising schedule $\bar{\alpha}_t$. 

\subsection{Model design}
Our denoising Diffusion model employs a U-Net architecture, as introduced by the probabilistic diffusion model in \cite{ho2020denoising}. To merge visual and semantic information effectively, we have customized this architecture to support both our time-dependent and visual-dependent embeddings, as illustrated in Fig. ~\ref{fig:second}. To our knowledge, this represents the first application of a U-Net architecture tailored for zero-shot learning in such a specific way. 
The encoder-decoder structure of our U-Net is composed of linear blocks featuring ReLU non-linearity and batch normalization. Inputs to each layer include sinusoidal time embeddings and conditional data, which are extracted using self-attention mechanisms and augmented by a skip-connection between the encoding and decoding stages.

\begin{algorithm}[t]
\caption{Training algorithm for RevCD}\label{alg:train}
\begin{algorithmic}

\Ensure
\State $ \bar{t} \sim TE(\mathcal{U}[0,1],d)$
\State $ \epsilon \sim \mathcal{N}(0, \mathbf{I}) $
    
\For {$s, \mathbf{x} \sim p(x,s) \in \mathcal{D}^{seen}$}

\State $\mathbf{x} \leftarrow \emptyset \textit{ with } p_{conditional}$
\State $s_0 \gets 2s-1$
\State $\hat{s}_0 \gets || Unet(\sqrt{\bar{\alpha_t}}s_{0}+\sqrt{(1-\bar{\alpha})\epsilon}, \bar{t}, x) - \sqrt{\bar{\alpha_t}}s_{0}+\sqrt{(1-\bar{\alpha})\epsilon}||_2^2$
\State $\hat{y} \gets \mathbb{E}_{\vartheta}(\hat{s}_0|y_{seen})$
\State $\hat{\epsilon}_t \gets ||\epsilon_0 - (\hat{s}_0-\epsilon_t)||_2^2$
\EndFor
\State $\text{Gradient step on:}$
\State $\nabla_{\theta}\ [\lambda_1 \mathcal{L}_{Diff}(\hat{s}_0) + \lambda_2 \mathcal{L}_{cls}(\hat{y}) + \lambda_3 \mathcal{L}_{noise} (\hat{\epsilon}_t)]$

\end{algorithmic}
\end{algorithm}


\subsection{Sampling}
Using standard methods from diffusion theory \cite{ho2020denoising}, we generate the semantic embedding space of an unseen sample through iterative conditional denoising using our trained model, as shown in Algorithm (~\ref{alg:sample}). Samples are drawn from the standard normal prior $p(s_T)\sim \mathcal{N}(0, \mathbf{I})$ and denoised conditioned on the sinusoidal time-step embedding $\bar{t}_i\ \forall i \in [1000,0]$, and the Transformer-encoded latent visual space $x \in \mathcal{D}^{unseen}$. 

The sampling through the reversed diffusion process is crucial for synthesizing high-quality semantic embeddings from the noised data. This process is governed by the following equation:

\begin{equation}
    s_{t-1} = \underbrace{\frac{1}{\sqrt{\alpha_t}}(s_t - \frac{(1-\alpha_t)\hat{s}_t}{\sqrt{1-\bar{\alpha}_t}})}_{\text{remove noise}} + \underbrace{\beta_t z}_{\text{add noise}}
\end{equation}


Here, \( s_{t-1} \) denotes the noisy semantic embeddings at time step \( t-1 \), \( \hat{s}_t \) represents the (predicted) noised sample at previous time step $t$, and \( \beta_t \) is the variance noise vector that controls the amount of noise added back to ensure stability, where $z \sim \mathcal{N}(0,\mathbf{}{I})$. This iterative refinement process enables the model to generate \( \hat{s}^{unseen}\) during inference.

\subsubsection{Classifier-free guidance.}
To improve semantic alignment during sampling, we adopt the classifier-free guidance framework \cite{ho2022classifier}. Using the difference between conditional and unconditional diffusion kernel estimates, we effectively steer the reverse process toward more probable samples under a desired condition.
Using  Tweedie's formula \cite{song2020score}, the denoising model can approximate the conditional and unconditional score functions $s_\theta(s_t, t, x) \approx \nabla_{s_t} \log p(s_t | x)$ $s_\theta(s_t, t, \emptyset) \approx \nabla_{s_t} \log p(s_t)$.

By leveraging the decomposition of the conditional score we can mirror a kind of gradient in the semantic space, pulling the diffusion kernel in a direction that increases likelihood under the condition.
\begin{equation}
\nabla_{s_t} \log p(s_t | x) = \nabla_{s_t} \log p(s_t) + \nabla_{s_t} \log p(x | s_t),
\end{equation}
where $\nabla_{s_t} \log p(x | s_t)$ acts as a classifier-like guidance term and can be reinterpreted as the difference between the conditional and unconditional scores:
\begin{equation}
\nabla_{s_t} \log p(x | s_t) \approx s_\theta(s_t, t, x) - s_\theta(s_t, t, \emptyset).
\end{equation}

To control the influence of the condition $x$, we scale this difference by a factor $g \in \mathbb{R}^+$, leading to the guided score estimate:
\begin{equation}
s_{\theta}(s_t, t, x) = s_{\theta}(s_t, t, \emptyset) + g \cdot \left[ s_\theta(s_t, t, x) - s_\theta(s_t, t, \emptyset) \right].
\end{equation}
This can be rearranged as our reversed transition function.
\begin{equation}
s_{\theta}(s_t, t, x) = (1 + g) \cdot s_{\theta}(s_t, t, x) - g \cdot s_\theta(s_t, t, \emptyset).
\end{equation}

Within the zero-shot learning framework, our goal is to reposition the probability mass of each conditional sample to enhance semantic separation, thereby improving recognition of unseen classes. To this end, we employ a single network jointly during sampling. The joint objective is illustrated as loss \raisebox{.5pt}{\textcircled{\raisebox{-.9pt} {3}}} in Fig.~\ref{fig:first}.

\begin{algorithm}[t]
\caption{Unseen sampling algorithm for RevCD}\label{alg:sample}
\begin{algorithmic}

\Ensure

\State $ s_t \sim \mathcal{N}(0, \mathbf{I}),\ x \sim p(x) \in \mathcal{D}^{unseen},\ g: \textit{guidance strength} $
\For {$t = T, ..., 1$}
\State $\bar{t} \gets TE(t,d)$
\State $s_t^c, s_t^u \gets Unet(s_t,\bar{t}, x),\ Unet(s_t,\bar{t}, \emptyset)$
\State $\hat{s} \gets (1 + g)s_t^c - gs_t^u$
\State $s_{t-1} \gets \frac{1}{\sqrt{\alpha_t}}(s_t - \frac{(1-\alpha_t)\hat{s}_t}{\sqrt{1-\bar{\alpha}_t}}) + \beta_t z$
\EndFor
\State \Return $(\hat{s}_0+1)\cdot\frac{1}{2} \quad \quad  \textit{\small{\#affine unmapping}}$ 
\end{algorithmic}
\end{algorithm}

\subsection{Zero-Shot Inference}
In the zero-shot learning setting, the model utilizes the denoising model $s_{\theta}$ to approximate the semantic distribution given the instance $x^{unseen}$. A pseudo sample drawn from this distribution is then classified using a nearest-neighbor approach in the semantic space, leveraging the semantic density to bridge the gap between visual features of $x^{unseen}$ and class labels $y^{unseen}$:

\begin{equation}
    \hat{y} = \arg\min_{y \in \mathcal{Y}_{unseen}} \text{dist}(\hat{s}^{unseen}, s^{unseen}_{y}),
\end{equation}

where \( \hat{y} \) is the predicted class label for an unseen class instance, and \( \text{dist}(\cdot, \cdot) \) denotes the distance metric, in our case cosine similarity:

\begin{equation}
    \text{dist(\textit{i,j})}= 1-\frac{\langle s_i, s_j \rangle}{||s_i||_2||s_j||_2}
    \label{eq:dist}
\end{equation}


\section{Experimental results}

We evaluate our approach by measuring classification accuracy on both known and unknown categories. Importantly, samples from unknown categories are entirely absent during training, ensuring that classification accuracy for these categories reflects the model's ability to transfer knowledge from the known space. This evaluation methodology aligns with established practices in zero-shot inference research, facilitating fair comparison and assessment of our model's performance.

\subsubsection{Dataset.} Our analysis of diffusion as a generative method for zero-shot inference employs four publicly available benchmark datasets, each widely used in the field. This allows us to make a fair comparison of the quality and semantic coverage of the approximations. The benchmark datasets used are: Animals with Attributes 2 dataset (AwA2) \cite{xian2017zero}, Caltech Birds dataset (CUB200-2011) \cite{wah2011caltech}, and Scene Understanding Attribute dataset (SUN) \cite{patterson2012sun}. 

The CUB dataset, focused on bird species, offers detailed representations in both image and semantic spaces. The semantic space consists of detailed attribute descriptions of the bird species’ characteristics. In contrast, AwA, which covers a range of different animal species, provides much coarser semantic descriptions, representing higher-level visual features. SUN consists of a scenery dataset with a wide range of classes, and its semantic descriptions are based on word2vec representations of each scene.

Visual features are derived using a ResNet101 backbone pre-trained on ImageNet \cite{xian2018zero}. We only compare models using similar image features to ensure a fair evaluation. We use the semantic attributes released with each dataset, which are derived from either crowd-sourced human annotations or word2vec-based label extractions.

\subsubsection{Implementation details.} The employed U-net architecture for our Diffusion model consists of three hidden, fully connected dense layers, ReLU activation functions, and dropout for regularization. We use a feature extractor with a multi-head self-attention layer (MSA) for the conditional space. In the encoder and the decoder of the U-net, we concatenate and add the sinusoidal time embedding to layer inputs and the conditional features as explained in section \ref{cross-had}. For our loss function, we fix $\lambda_1=1$ and $\lambda_2=1$ during training, while $\lambda_3$ varies depending on the dataset (see section \ref{sec:cls_loss}).

\subsection{Generalized accuracy}
Our method achieves strong performance in the generalized zero-shot learning (GZSL) setting, particularly excelling on seen (S) class classification across all datasets. See Table ~\ref{res:main}. On AWA, we achieve the highest seen accuracy of 94.5\%, substantially outperforming the previous best. Similarly, for CUB and SUN, we report 87.5\% and 66.9\% on seen classes, again outperforming existing state-of-the-art methods. We attribute this to the conditional nature of our diffusion model. Unlike prior approaches that condition on semantics to generate visuals, our reversed formulation conditions on visual inputs, allowing the model to preserve fine-grained visual distinctions. This becomes especially beneficial in datasets like AWA and SUN, where visual cues are rich and diverse. This highlights the advantage of diffusion models in capturing complex visual semantics through iterative refinement.

\begin{table}[h]
\caption{Generalized zero-shot learning (GZSL) results on AWA, CUB, and SUN datasets. S, U, and H denote accuracy on seen classes, unseen classes, and their harmonic mean, respectively. Our method achieves the highest seen class accuracy across all datasets (such as 94.5\% on AWA, 87.5\% on CUB), and a competitive harmonic mean, especially on SUN (52.6\%), highlighting the effectiveness of our conditional diffusion formulation.}
\begin{tabular}{@{}rc|ccc|ccc|ccc@{}}
\toprule
                &             & \multicolumn{3}{c|}{AWA} & \multicolumn{3}{c|}{CUB} & \multicolumn{3}{c}{SUN} \\ \midrule
Model           & Venue       & S      & U      & H      & S      & U      & H      & S      & U      & H     \\ \midrule
f-VAEGAN-D2     & CVPR(19)    & 70.6   & 57.6   & 63.5   & 60.1   & 48.4   & 53.6   & 38.0   & 45.1   & 41.3  \\
LisGAN          & CVPR(19)    & 76.3   & 52.6   & 62.3   & 57.9   & 46.5   & 51.6   & 37.8   & 42.9   & 40.2  \\
GDAN            & ICCV(19)    & 67.5   & 32.1   & 43.5   & 66.7   & 39.3   & 49.5   & 40.9   & 38.1   & 53.4  \\
ZSML            & IAAA(20)    & 58.9   & 74.6   & 65.8   & 60.0   & 52.1   & 55.7   & 45.1   & 21.7   & 29.3  \\
DAZLE           & CVPR(20)    & 75.7   & 60.3   & 67.1   & 59.6   & 56.7   & 58.1   & 24.3   & 52.3   & 33.2  \\
SDGZSL          & ICCV(21)    & 73.6   & 64.6   & 68.8   & 66.4   & 59.9   & 63.0   & -      & -      & -     \\
DAA-ZSL         & --          & 79.9   & 65.7   & 72.1   & 65.5   & \textbf{66.1}   & 65.8   & 38.7   & 47.8   & 42.8  \\
HSVA            & NeurIPS(21) & 76.6   & 53.9   & 66.8   & 58.3   & 57.2   & 55.3   & 39.0   & 48.6   & 43.3  \\
ICCE            & CVPR(22)    & 65.3   & \textbf{82.3}   & 72.8   & 65.5   & 67.3   & 66.4   & -      & -      & -     \\
FREE+ESZSL      & ICLR(22)    & 78.0   & 51.3   & 61.8   & 60.4   & 51.6   & 55.7   & 36.5   & 48.2   & 41.5  \\
TF-VAEGAN+ESZSL & ICLR(22)    & 74.7   & 55.2   & 63.5   & 63.3   & 51.1   & 56.6   & 39.7   & 44.0   & 41.7  \\
TransZero       & AAAI(22)    & 61.3   & 82.3   & 70.2   & 69.3   & 68.3   &\textbf{ 68.8}   & 52.6   & 33.4   & 40.8  \\
TDCSS           & CVPR(22)    & 59.2   & 74.9   & 66.1   & 44.2   & 62.8   & 51.9   & -      & -      & -     \\
ZLAP            & IJCAI(22)   & 76.3   & 74.7   & 75.5   & 32.4   & 25.5   & 28.5   & 48.1   & 47.2   & 47.7  \\
SE-GZSL         & AAAI(23)    & 68.1   & 58.3   & 62.8   & 53.3   & 41.5   & 46.7   & 30.5   & \textbf{50.9}   & 34.9  \\
TPR             & NeurIPS(24) & 87.1   & 76.8   & 81.6   & 41.2   & 26.8   & 32.5   & 50.4   & 45.4   & 47.8  \\
MAIN            & WACV(24)    & 81.8   & 72.1   & \textbf{76.7}   & 58.7   & 65.9   & 62.1   & 40.0   & 50.1   & 48.8  \\ \midrule
Ours            &             & \textbf{94.5}   & 42.4   & 58.3   & \textbf{87.5}   & 32.3   & 47.2   & \textbf{66.9}   & 43.4   & \textbf{52.6}  \\ \bottomrule
\end{tabular}
\label{res:main}
\end{table}

However, our approach slightly underperforms on unseen (U) classes compared to some state-of-the-art methods. For instance, on AWA, we report 42.4\%, compared to 82.3\% achieved by ICCE. This is a result of the semantic space being estimated from a single visual instance: since each visual sample may correspond to several valid semantic embeddings, the diffusion model may distribute probability mass across divergent regions of the semantic space. This inherent multi-modality poses challenges in datasets with coarse or ambiguous semantic descriptors (e.g., AWA). Notably, our method performs more competitively on SUN’s unseen classes (43.4\%, close to the best-performing method at 50.9\%), where the broad and distributed semantic space (based on word2vec scene embeddings) mitigates this effect. Overall, our method achieves competitive harmonic means across all datasets: 58.3\% (AWA), 47.2\% (CUB), and 52.6\% (SUN), while showcasing the strength of diffusion-based conditional modeling for generalized classification tasks.

\subsection{Posterior approximation}
We demonstrate semantic posterior sampling using our Diffusion model. To evaluate its performance, we consider two natural comparisons. (i) Models that use variational inference to approximate the posterior, such as VAEs, which are optimized by balancing reconstruction accuracy and the divergence between the approximate and true posterior distributions; and (ii) models that use indirect approaches to approximate the distribution, such as GANs, which achieve posterior matching through adversarial training.

\begin{table*}[h]
\centering
\caption{Result of generalized ZSL for classification, for the most prominent generative approaches. \textsuperscript{\textdagger} denotes the model consists of additional components that are disregarded.
}
\begin{adjustbox}{width=1\textwidth,center=\textwidth}
\begin{tabular}{@{}c|l|ccc|ccc|ccc|@{}}
\toprule
\begin{tabular}[c]{@{}c@{}}Generative \\ model\end{tabular}      & Name  & \multicolumn{3}{c|}{AwA}                      & \multicolumn{3}{c|}{CUB}                      & \multicolumn{3}{c|}{SUN}                      \\ \midrule
                                                                 &       & Seen          & Unseen        & Harm.         & Seen          & Unseen        & Harm.         & Seen          & Unseen        & Harm.         \\ \cmidrule(l){2-11} 
VAE\textsuperscript\textdagger\ & cVAE  & 72.6          & \textbf{54.4} & \textbf{62.2} & 59.9          & \textbf{47.0} & \textbf{52.7} & -             & -             & -             \\
GAN\textsuperscript\textdagger & GAN   & 82.4 & 24.7          & 38.1          & 44.4          & 31.3          & 36.8          & 43.3          & 29.0          & 31.4          \\ \midrule
Diffusion (ours)                      & RevCD & \textbf{94.5} & 42.4          & 58.3          & \textbf{87.5} & 32.3          & 47.2          & \textbf{66.9} & \textbf{43.4} & \textbf{52.6} \\ \bottomrule
\end{tabular}
\end{adjustbox}
\label{res:generative}
\end{table*}

These comparisons are summarized in Table \ref{res:generative}. As shown, no single generative model consistently outperforms the others across all datasets when generating both seen and unseen samples. Notably, our method surpasses the other approaches in generating samples when measured by the harmonic mean. The most significant performance gap is observed in the semantically coarse SUN dataset, where our approach achieves a 20\% improvement over GANs. A similar trend is evident in the class-diverse AWA dataset, albeit with smaller margins. In contrast, the CUB dataset, which features a wide variety of fine-grained semantic details, proves challenging for denoising approaches, making variational inference methods a more effective fit.

VAEs benefit from the tractable estimation of the posterior distribution, as evidenced by a 5.4\% higher harmonic mean when both seen and unseen samples are drawn from tighter distributions, such as those observed in the CUB dataset, which emphasizes local descriptions. In contrast, GANs may underperform in this context, likely due to mode collapse in the posterior. Our Diffusion model performs moderately, achieving a 10.4\% improvement over GANs.

Conversely, GANs implicitly learn the distribution through adversarial training, which encourages the generator to produce high-fidelity samples, as demonstrated in the AWA dataset, where attributes emphasize global image descriptions. However, the absence of explicit density estimation makes GANs susceptible to seen-unseen bias, leading to a preference for the seen distribution during inference, as observed in the SUN datasets. Our Diffusion model shows strong performance in generating samples when the seen distribution is discriminative and low-dimensional, as in the AWA. However, it struggles to maintain a tight lower bound on the true data distribution as the dimensionality of the semantic space increases, as evidenced by its performance in the CUB dataset.

\begin{figure}[h]
\centering
\begin{subfigure}[t]{0.5\linewidth}
    \centering
    \includegraphics[width=\linewidth]{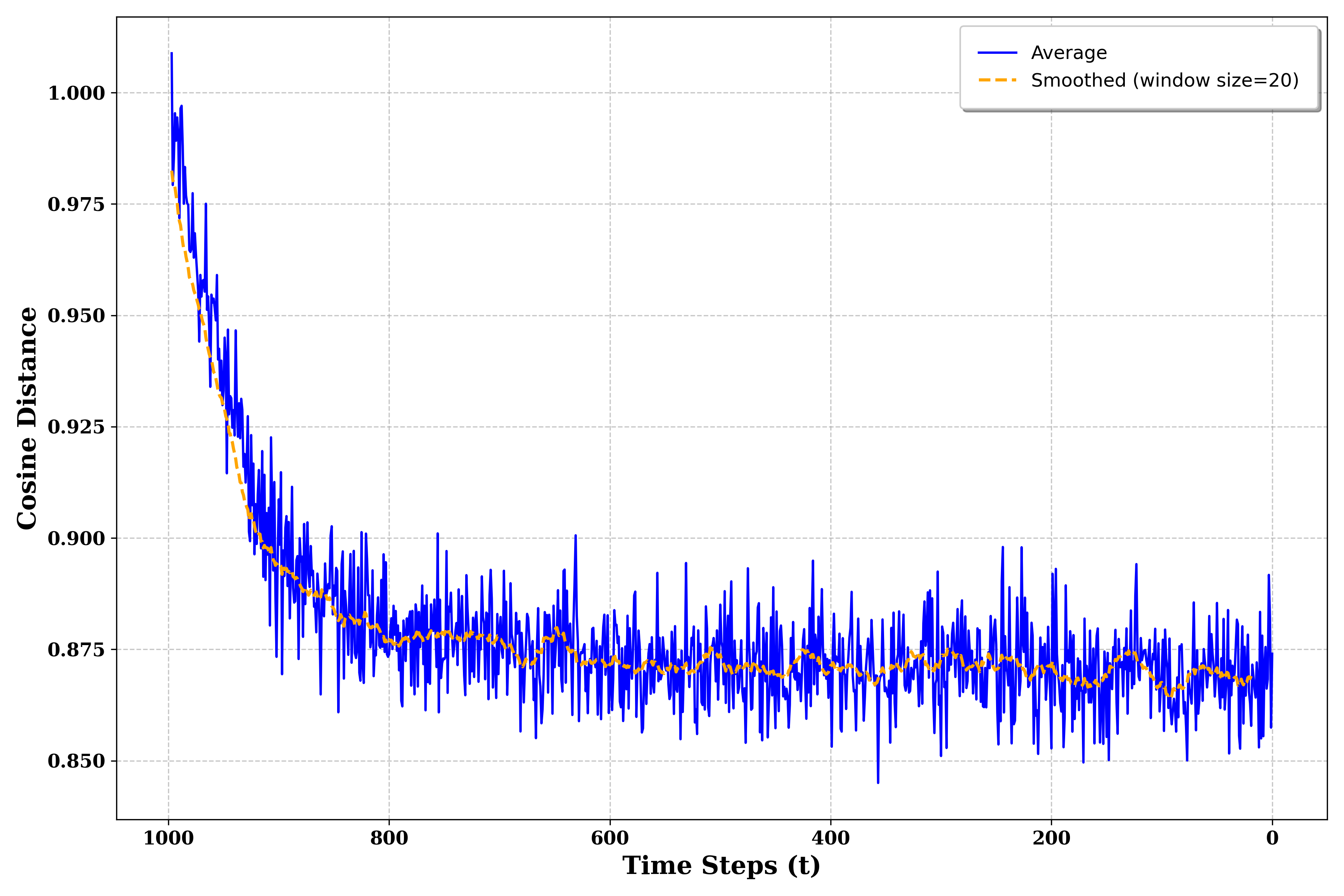}
    \caption{}
\end{subfigure}\hfil
\begin{subfigure}[t]{0.5\linewidth}
    \centering
    \includegraphics[width=\linewidth]{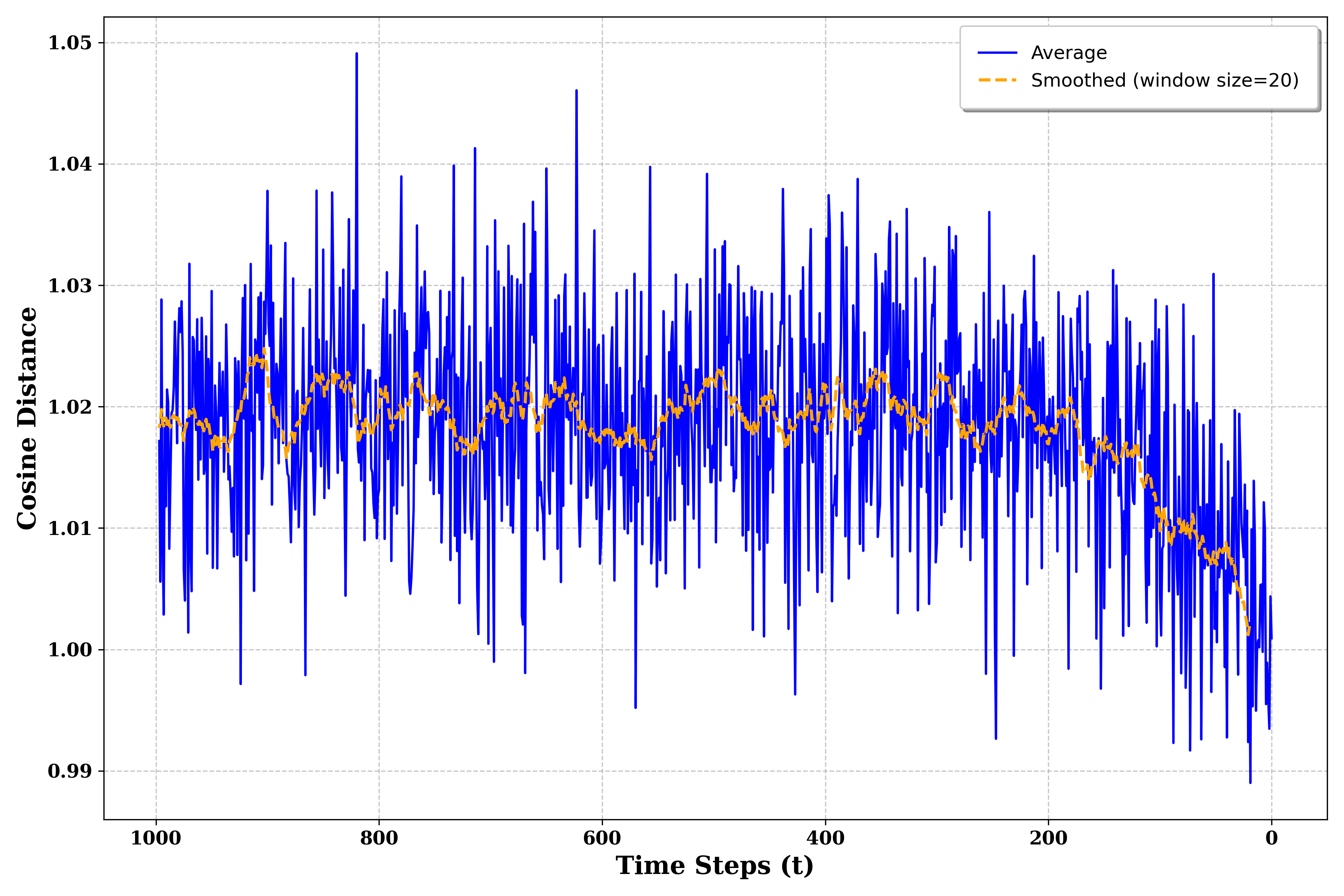}
    \caption[b]{}
 \end{subfigure}
 \caption{The cosine distance to the true semantic space and the denoised learned representation are shown for both the AWA dataset and the CUB dataset. \textbf{(a)} For AWA, we observed a rapid reduction in noise in the initial timesteps, which gradually slowed as it approached the fully denoised space. \textbf{(b)} Conversely, for the CUB dataset, which possesses a semantically fine-grained space, the reduction in noise exhibited an inverse pattern.}
\label{fig:sem_dist}
\end{figure}

This pattern is evident in Fig. \ref{fig:sem_dist}, which illustrates the sample quality during iterative denoising. In Fig. \ref{fig:sem_dist}(a), the density of seen samples in the AWA dataset is reproduced more quickly compared to the higher-dimensional space in the CUB dataset, as shown in Fig. \ref{fig:sem_dist}(b).

\subsection{Effect of classification loss} \label{sec:cls_loss}
\begin{figure}[h]
    \centering
    \begin{subfigure}[t]{0.32\linewidth}
        \includegraphics[width=\linewidth]{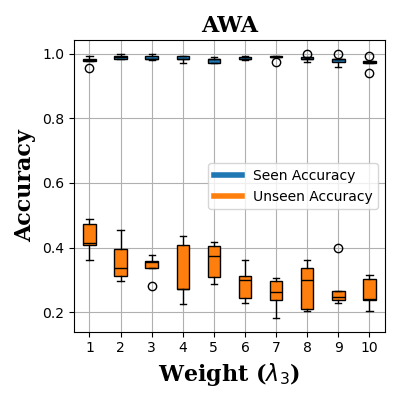}
        \caption{}
    \end{subfigure}
    \hfill
    \begin{subfigure}[t]{0.32\linewidth}
        \includegraphics[width=\linewidth]{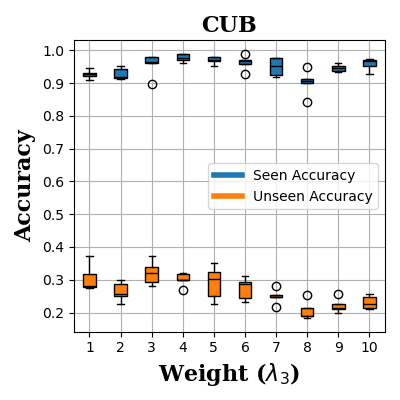}
        \caption{}
    \end{subfigure}
    \hfill
    \begin{subfigure}[t]{0.32\linewidth}
        \includegraphics[width=\linewidth]{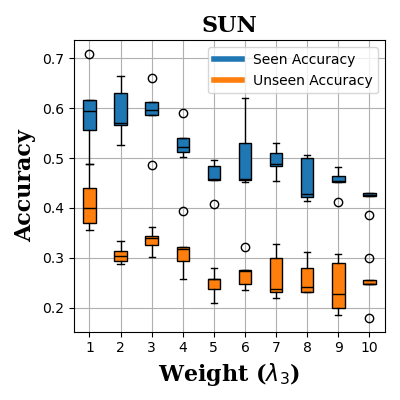}
        \caption{}
    \end{subfigure}
    \caption{Seen and unseen accuracy as a function of the classification loss weight $\lambda_3$ across AWA, CUB, and SUN. Increasing $\lambda_3$ consistently improves seen class accuracy but degrades unseen performance, with the trade-off varying by dataset. AWA exhibits sharp overfitting due to coarse semantics, while CUB remains more stable under stronger supervision. SUN shows a more balanced response, reflecting its broader semantic diversity.}
    \label{fig:cls_weigth}
\end{figure}

We investigate the influence of the classification loss weight $\lambda_3$ on the performance of our diffusion-based model (see Fig. (\ref{fig:cls_weigth})). The classification term guides the forward process during kernel estimation by biasing the diffusion trajectory toward semantically dense regions. Specifically, it introduces a gradient-based constraint that guides the process toward class-consistent subspaces, refining the learned transition kernel to denoise representations aligned with class identity.

Our reversed process estimates the semantic density of each class. The effect of this weight is notably different across datasets. On AWA, where semantic representations are coarse and broadly defined, increased weighting of the classification loss substantially boosts seen class accuracy—reaching near-perfect performance across all values of $\lambda_3$. However, this comes at the expense of unseen accuracy, which decreases as the model overfits to seen class modes, likely due to the reduced specificity in the semantic space, Fig. (\ref{fig:cls_weigth}a). In contrast, CUB, characterized by fine-grained and richly structured semantic attributes, maintains high seen accuracy even at moderate classification weights, with a more stable and gradual decline in unseen performance, Fig. (\ref{fig:cls_weigth}b). SUN, which features a broader class diversity and distributed word2vec-based semantics, displays a well-balanced trade-off. Seen accuracy is maximized at lower $\lambda_3$ values, while unseen performance remains relatively stable before degrading at higher weights, Fig. (\ref{fig:cls_weigth}c). The robustness observed in SUN can be attributed to the high semantic variability, which naturally regularizes the kernel estimation and prevents premature convergence toward over-specialized representations.

These results highlight that the optimal weight of the classification-guided diffusion kernel is dataset-dependent. Fine-grained datasets like CUB benefit from moderate supervision, while coarsely defined or semantically diverse datasets (AWA and SUN) require careful calibration to avoid collapsing the semantic diversity necessary for zero-shot generalization.

\section{Conclusion}
In this paper, we introduce a reversed Conditional Diffusion model (RevCD) and evaluate its performance against state-of-the-art methods for generalized zero-shot learning. Additionally, by comparing with VAEs and GANs, our research explores the largely untapped potential of diffusion-based models to generate unseen samples. Our RevCD model estimates the semantic density for classes, which is then used to generate prototypes of both seen and unseen instances for high-accuracy classification. By leveraging visual conditioning, our approach enables precise control over the generation process and improved posterior approximations, outperforming other generative methods in settings with seen classes. Experimental results demonstrate the advantages of using a diffusion model as a generative backbone, especially regarding its robustness to diverse semantic information. We believe our findings can stimulate further exploration of diffusion models in generalized zero-shot learning (GZSL). Moreover, expanding cross-dataset evaluations in future zero-shot learning research could lead to the development of more resilient models.


%

\bibliographystyle{splncs04}
\bibliography{mybibliography}

\begin{thebibliography}{10}
\providecommand{\url}[1]{\texttt{#1}}
\providecommand{\urlprefix}{URL }
\providecommand{\doi}[1]{https://doi.org/#1}

\bibitem{akata2015label}
Akata, Z., Perronnin, F., Harchaoui, Z., Schmid, C.: Label-embedding for image classification. IEEE transactions on pattern analysis and machine intelligence  \textbf{38}(7),  1425--1438 (2015)

\bibitem{alamri2023implicit}
Alamri, F., Dutta, A.: Implicit and explicit attention mechanisms for zero-shot learning. Neurocomputing  \textbf{534},  55--66 (2023)

\bibitem{azizi2023synthetic}
Azizi, S., Kornblith, S., Saharia, C., Norouzi, M., Fleet, D.J.: Synthetic data from diffusion models improves imagenet classification. arXiv preprint arXiv:2304.08466  (2023)

\bibitem{bau2019seeing}
Bau, D., Zhu, J.Y., Wulff, J., Peebles, W., Strobelt, H., Zhou, B., Torralba, A.: Seeing what a gan cannot generate. In: Proceedings of the IEEE/CVF international conference on computer vision. pp. 4502--4511 (2019)

\bibitem{bucher2017generating}
Bucher, M., Herbin, S., Jurie, F.: Generating visual representations for zero-shot classification. In: Proceedings of the IEEE International Conference on Computer Vision Workshops. pp. 2666--2673 (2017)

\bibitem{chen2023robust}
Chen, H., Dong, Y., Wang, Z., Yang, X., Duan, C., Su, H., Zhu, J.: Robust classification via a single diffusion model. arXiv preprint arXiv:2305.15241  (2023)

\bibitem{chen2021semantics}
Chen, Z., Luo, Y., Qiu, R., Wang, S., Huang, Z., Li, J., Zhang, Z.: Semantics disentangling for generalized zero-shot learning. In: Proceedings of the IEEE/CVF international conference on computer vision. pp. 8712--8720 (2021)

\bibitem{chen2020rethinking}
Chen, Z., Wang, S., Li, J., Huang, Z.: Rethinking generative zero-shot learning: An ensemble learning perspective for recognising visual patches. In: Proceedings of the 28th ACM International Conference on Multimedia. pp. 3413--3421 (2020)

\bibitem{clark2024text}
Clark, K., Jaini, P.: Text-to-image diffusion models are zero shot classifiers. Advances in Neural Information Processing Systems  \textbf{36} (2024)

\bibitem{ding2023enhanced}
Ding, B., Fan, Y., He, Y., Zhao, J.: Enhanced vaegan: a zero-shot image classification method. Applied Intelligence  \textbf{53}(8),  9235--9246 (2023)

\bibitem{duchi2007derivations}
Duchi, J.: Derivations for linear algebra and optimization. Berkeley, California  \textbf{3}(1),  2325--5870 (2007)

\bibitem{frome2013devise}
Frome, A., Corrado, G.S., Shlens, J., Bengio, S., Dean, J., Ranzato, M., Mikolov, T.: Devise: A deep visual-semantic embedding model. Advances in neural information processing systems  \textbf{26} (2013)

\bibitem{gao2020zero}
Gao, R., Hou, X., Qin, J., Chen, J., Liu, L., Zhu, F., Zhang, Z., Shao, L.: Zero-vae-gan: Generating unseen features for generalized and transductive zero-shot learning. IEEE Transactions on Image Processing  \textbf{29},  3665--3680 (2020)

\bibitem{goodfellow2014generative}
Goodfellow, I., Pouget-Abadie, J., Mirza, M., Xu, B., Warde-Farley, D., Ozair, S., Courville, A., Bengio, Y.: Generative adversarial nets. Advances in neural information processing systems  \textbf{27} (2014)

\bibitem{gupta2023generative}
Gupta, A., Narayan, S., Khan, S., Khan, F.S., Shao, L., van~de Weijer, J.: Generative multi-label zero-shot learning. IEEE Transactions on Pattern Analysis and Machine Intelligence  (2023)

\bibitem{han2021contrastive}
Han, Z., Fu, Z., Chen, S., Yang, J.: Contrastive embedding for generalized zero-shot learning. In: Proceedings of the IEEE/CVF conference on computer vision and pattern recognition. pp. 2371--2381 (2021)

\bibitem{hao2023learning}
Hao, S., Han, K., Wong, K.Y.K.: Learning attention as disentangler for compositional zero-shot learning. In: Proceedings of the IEEE/CVF Conference on Computer Vision and Pattern Recognition. pp. 15315--15324 (2023)

\bibitem{heyden2023integral}
Heyden, W., Ullah, H., Siddiqui, M.S., Al~Machot, F.: An integral projection-based semantic autoencoder for zero-shot learning. IEEE Access  (2023)

\bibitem{ho2020denoising}
Ho, J., Jain, A., Abbeel, P.: Denoising diffusion probabilistic models. Advances in neural information processing systems  \textbf{33},  6840--6851 (2020)

\bibitem{ho2022classifier}
Ho, J., Salimans, T.: Classifier-free diffusion guidance. arXiv preprint arXiv:2207.12598  (2022)

\bibitem{huang2019generative}
Huang, H., Wang, C., Yu, P.S., Wang, C.D.: Generative dual adversarial network for generalized zero-shot learning. In: Proceedings of the IEEE/CVF conference on computer vision and pattern recognition. pp. 801--810 (2019)

\bibitem{ji2023zero}
Ji, Z., Cui, B., Yu, Y., Pang, Y., Zhang, Z.: Zero-shot classification with unseen prototype learning. Neural computing and applications pp. 1--11 (2023)

\bibitem{khan2023learning}
Khan, M.G.Z.A., Naeem, M.F., Van~Gool, L., Pagani, A., Stricker, D., Afzal, M.Z.: Learning attention propagation for compositional zero-shot learning. In: Proceedings of the IEEE/CVF Winter Conference on Applications of Computer Vision. pp. 3828--3837 (2023)

\bibitem{kingma2013auto}
Kingma, D.P., Welling, M.: Auto-encoding variational bayes. arXiv preprint arXiv:1312.6114  (2013)

\bibitem{lampert2009learning}
Lampert, C.H., Nickisch, H., Harmeling, S.: Learning to detect unseen object classes by between-class attribute transfer. In: 2009 IEEE conference on computer vision and pattern recognition. pp. 951--958. IEEE (2009)

\bibitem{lazaridou2015hubness}
Lazaridou, A., Dinu, G., Baroni, M.: Hubness and pollution: Delving into cross-space mapping for zero-shot learning. In: Zong C, Strube M, editors. Proceedings of the 53rd Annual Meeting of the Association for Computational Linguistics and the 7th International Joint Conference on Natural Language Processing (Volume 1: Long Papers); 2015 Jul 26-31; Beijing, China. Stroudsburg (PA): Association for Computational Linguistics; 2015. p. 270-80. ACL (Association for Computational Linguistics) (2015)

\bibitem{li2023your}
Li, A.C., Prabhudesai, M., Duggal, S., Brown, E., Pathak, D.: Your diffusion model is secretly a zero-shot classifier. In: Proceedings of the IEEE/CVF International Conference on Computer Vision. pp. 2206--2217 (2023)

\bibitem{li2018visualizing}
Li, H., Xu, Z., Taylor, G., Studer, C., Goldstein, T.: Visualizing the loss landscape of neural nets. Advances in neural information processing systems  \textbf{31} (2018)

\bibitem{li2023distilled}
Li, Y., Liu, Z., Jha, S., Yao, L.: Distilled reverse attention network for open-world compositional zero-shot learning. In: Proceedings of the IEEE/CVF International Conference on Computer Vision. pp. 1782--1791 (2023)

\bibitem{liu2024transductive}
Liu, Y., Tao, K., Tian, T., Gao, X., Han, J., Shao, L.: Transductive zero-shot learning with generative model-driven structure alignment. Pattern Recognition p. 110561 (2024)

\bibitem{lu2022learn}
Lu, Z., Lu, Z., Yu, Y., Wang, Z.: Learn more from less: Generalized zero-shot learning with severely limited labeled data. Neurocomputing  (2022)

\bibitem{lucas2019don}
Lucas, J., Tucker, G., Grosse, R.B., Norouzi, M.: Don't blame the elbo! a linear vae perspective on posterior collapse. Advances in Neural Information Processing Systems  \textbf{32} (2019)

\bibitem{luo2022understanding}
Luo, C.: Understanding diffusion models: A unified perspective. arXiv preprint arXiv:2208.11970  (2022)

\bibitem{mishra2018generative}
Mishra, A., Krishna~Reddy, S., Mittal, A., Murthy, H.A.: A generative model for zero shot learning using conditional variational autoencoders. In: Proceedings of the IEEE conference on computer vision and pattern recognition workshops. pp. 2188--2196 (2018)

\bibitem{patterson2012sun}
Patterson, G., Hays, J.: Sun attribute database: Discovering, annotating, and recognizing scene attributes. In: 2012 IEEE Conference on Computer Vision and Pattern Recognition. pp. 2751--2758. IEEE (2012)

\bibitem{pourpanah2022review}
Pourpanah, F., Abdar, M., Luo, Y., Zhou, X., Wang, R., Lim, C.P., Wang, X.Z., Wu, Q.J.: A review of generalized zero-shot learning methods. IEEE transactions on pattern analysis and machine intelligence  \textbf{45}(4),  4051--4070 (2022)

\bibitem{salimans2016improved}
Salimans, T., Goodfellow, I., Zaremba, W., Cheung, V., Radford, A., Chen, X.: Improved techniques for training gans. Advances in neural information processing systems  \textbf{29} (2016)

\bibitem{schonfeld2019generalized}
Schonfeld, E., Ebrahimi, S., Sinha, S., Darrell, T., Akata, Z.: Generalized zero-and few-shot learning via aligned variational autoencoders. In: Proceedings of the IEEE/CVF conference on computer vision and pattern recognition. pp. 8247--8255 (2019)

\bibitem{shipard2023diversity}
Shipard, J., Wiliem, A., Thanh, K.N., Xiang, W., Fookes, C.: Diversity is definitely needed: Improving model-agnostic zero-shot classification via stable diffusion. In: Proceedings of the IEEE/CVF Conference on Computer Vision and Pattern Recognition. pp. 769--778 (2023)

\bibitem{socher2013zero}
Socher, R., Ganjoo, M., Manning, C.D., Ng, A.: Zero-shot learning through cross-modal transfer. Advances in neural information processing systems  \textbf{26} (2013)

\bibitem{sohl2015deep}
Sohl-Dickstein, J., Weiss, E., Maheswaranathan, N., Ganguli, S.: Deep unsupervised learning using nonequilibrium thermodynamics. In: International conference on machine learning. pp. 2256--2265. PMLR (2015)

\bibitem{song2020score}
Song, Y., Sohl-Dickstein, J., Kingma, D.P., Kumar, A., Ermon, S., Poole, B.: Score-based generative modeling through stochastic differential equations. arXiv preprint arXiv:2011.13456  (2020)

\bibitem{vaswani2017attention}
Vaswani, A., Shazeer, N., Parmar, N., Uszkoreit, J., Jones, L., Gomez, A.N., Kaiser, {\L}., Polosukhin, I.: Attention is all you need. Advances in neural information processing systems  \textbf{30} (2017)

\bibitem{verma2017simple}
Verma, V.K., Rai, P.: A simple exponential family framework for zero-shot learning. In: Machine Learning and Knowledge Discovery in Databases: European Conference, ECML PKDD 2017, Skopje, Macedonia, September 18--22, 2017, Proceedings, Part II 10. pp. 792--808. Springer (2017)

\bibitem{wah2011caltech}
Wah, C., Branson, S., Welinder, P., Perona, P., Belongie, S.: The caltech-ucsd birds-200-2011 dataset  (2011)

\bibitem{wang2023generalized}
Wang, Q., Breckon, T.P.: Generalized zero-shot domain adaptation via coupled conditional variational autoencoders. Neural Networks  \textbf{163},  40--52 (2023)

\bibitem{wang2023zero}
Wang, Y., Feng, L., Song, X., Xu, D., Zhai, Y.: Zero-shot image classification method based on attention mechanism and semantic information fusion. Sensors  \textbf{23}(4), ~2311 (2023)

\bibitem{wang2015semi}
Wang, Y., Chen, S., Xue, H., Fu, Z.: Semi-supervised classification learning by discrimination-aware manifold regularization. Neurocomputing  \textbf{147},  299--306 (2015)

\bibitem{xian2016latent}
Xian, Y., Akata, Z., Sharma, G., Nguyen, Q., Hein, M., Schiele, B.: Latent embeddings for zero-shot classification. In: Proceedings of the IEEE conference on computer vision and pattern recognition. pp. 69--77 (2016)

\bibitem{xian2018zero}
Xian, Y., Lampert, C.H., Schiele, B., Akata, Z.: Zero-shot learning—a comprehensive evaluation of the good, the bad and the ugly. IEEE transactions on pattern analysis and machine intelligence  \textbf{41}(9),  2251--2265 (2018)

\bibitem{xian2017zero}
Xian, Y., Schiele, B., Akata, Z.: Zero-shot learning-the good, the bad and the ugly. In: Proceedings of the IEEE Conference on Computer Vision and Pattern Recognition. pp. 4582--4591 (2017)

\bibitem{zhang2023data}
Zhang, J., Liao, S., Zhang, H., Long, Y., Zhang, Z., Liu, L.: Data driven recurrent generative adversarial network for generalized zero shot image classification. Information Sciences  \textbf{625},  536--552 (2023)

\end{thebibliography}

\end{document}